\documentclass[letterpaper, 10 pt, conference]{ieeeconf}  % Comment this line out if you need a4paper

\IEEEoverridecommandlockouts                              % This command is only needed if 
\usepackage{graphics} % for pdf, bitmapped graphics files
\usepackage{amsmath} % assumes amsmath package installed
\usepackage{amssymb}  % assumes amsmath package installed
\usepackage{multirow}
\usepackage{cleveref}
\usepackage{booktabs}
\usepackage[table,xcdraw]{xcolor}
\usepackage{arydshln}
\usepackage{graphicx}

\usepackage{xcolor}
\usepackage{etoolbox}
\newcommand{\ours}{{DSP}}

\newcommand{\ci}[1]{\scriptsize{~($\pm #1$)}}
\newcommand{\ourcell}{\cellcolor{gray!10}}

\title{\LARGE \bf
Diffusion Stabilizer Policy \\for Automated Surgical Robot Manipulations
}

\author{Chonlam Ho$^{1,*}$, Jianshu Hu$^{1,*}$, Lei Song$^{2}$, Hesheng Wang$^{3}$, Qi Dou$^{2}$, and Yutong Ban$^{1}$% <-this % stops a space
\thanks{$^{1}$ Chonlam Ho, Jianshu Hu and Yutong Ban are with UM-SJTU Joint Intuitute, Shanghai Jiao Tong University,
        Shanghai, China
        {\tt\small \{raphaelho2001, hjs1998, yban\}@sjtu.edu.cn}}%
\thanks{$^{2}$Lei Song and Qi Dou are with the Department of Computer Science and Engineering, The Chinese University of Hong Kong,
        HongKong, China
        {\tt\small qidou@cuhk.edu.hk}}%
\thanks{$^{3}$Hesheng Wang is with the Department of Automation, Shanghai Jiao Tong Univerisity,
        Shanghai, China
        {\tt\small wanghesheng@sjtu.edu.cn}}%
\thanks{*indicates the authors contributed equally}% 
\thanks{Corresponding email: {\tt\small  yban@sjtu.edu.cn}}%
}

\begin{document}

\maketitle
\thispagestyle{empty}
\pagestyle{empty}

%%%%%%%%%%%%%%%%%%%%%%%%%%%%%%%%%%%%%%%%%%%%%%%%%%%%%%%%%%%%%%%%%%%%%%%%%%%%%%%%
\begin{abstract}
Intelligent surgical robots have the potential to revolutionize clinical practice by enabling more precise and automated surgical procedures.
However, the automation of such robot for surgical tasks remains under-explored compared to recent advancements in solving household manipulation tasks.
These successes have been largely driven by (1) advanced models, such as transformers and diffusion models, and (2) large-scale data utilization.
Aiming to extend these successes to the domain of surgical robotics, we propose a diffusion-based policy learning framework, called Diffusion Stabilizer Policy (\ours), which enables training with imperfect, perturbed or even failed trajectories.
Our approach consists of two stages: first, we train the diffusion stabilizer policy using only clean data.
Then, the policy is continuously updated using a mixture of clean and perturbed data, with filtering based on the prediction error on actions.
Comprehensive experiments conducted in both simulation and real-world demonstrate the superior performance of our method under different types of perturbations.
% Supplementary videos intuitively showcasing the automation process are provided to accompany this work.
% Code will be released upon acceptance.
\end{abstract}

%%%%%%%%%%%%%%%%%%%%%%%%%%%%%%%%%%%%%%%%%%%%%%%%%%%%%%%%%%%%%%%%%%%%%%%%%%%%%%%%
\section{INTRODUCTION}

Surgical robots have the power of enhancing the capabilities of surgeons by offering greater dexterity and stability in finishing tasks such as suturing \cite{9551569, 8698220},
% \cite{9551569, 9223543, doi:10.1126/scitranslmed.aad9398, 8794306, 8698220, 7989278}
tissue manipulation \cite{doi:10.1126/scirobotics.abj2908, 9636175}, and endoscope control \cite{10.1109/COASE.2018.8560468, 8627918}.
% \cite{robotics3030310, 9895213, 10.1109/COASE.2018.8560468, 8627918, DBLP:journals/firai/GruijthuijsenGB22}
Furthermore, surgical robots facilitate remote surgeries, paving the way for telemedicine and enabling operation on patients from a distance.
The well-known da Vinci Surgical Research Kit (dVRK) system has been widely used in clinical applications.
Automation of such surgical robots is an essential step for advancing precision, promoting accessibility and reducing burden of surgeon in medical procedures.
However, the automation of such robots for surgical tasks is under-explored compared to recent advancement \cite{pi_0, rt-x} in solving household tasks using data-driven approaches.

Among data-driven policy learning methods, imitation learning and reinforcement learning have become two predominant approaches.
Powered by the ability of modeling multi-modal data distribution, diffusion policy \cite{diffusion_policy}
% \cite{decision_diffuser, diffusion_policy, 3D_diffusion_policy, consistency_diffusion_policy}
has become a significant branch of work in imitation learning.
However, it requires a large amount of data for generalization and its performance highly relies on data quality.
Given that imperfect demonstrations with perturbations in real-world data collection are somehow inevitable, properly leveraging perturbed and even failed data could possibly extend the successful application of data scaling in solving household manipulation tasks to the domain of surgical robots.

\begin{figure}[t]
\centering
\includegraphics[width=0.45\textwidth]{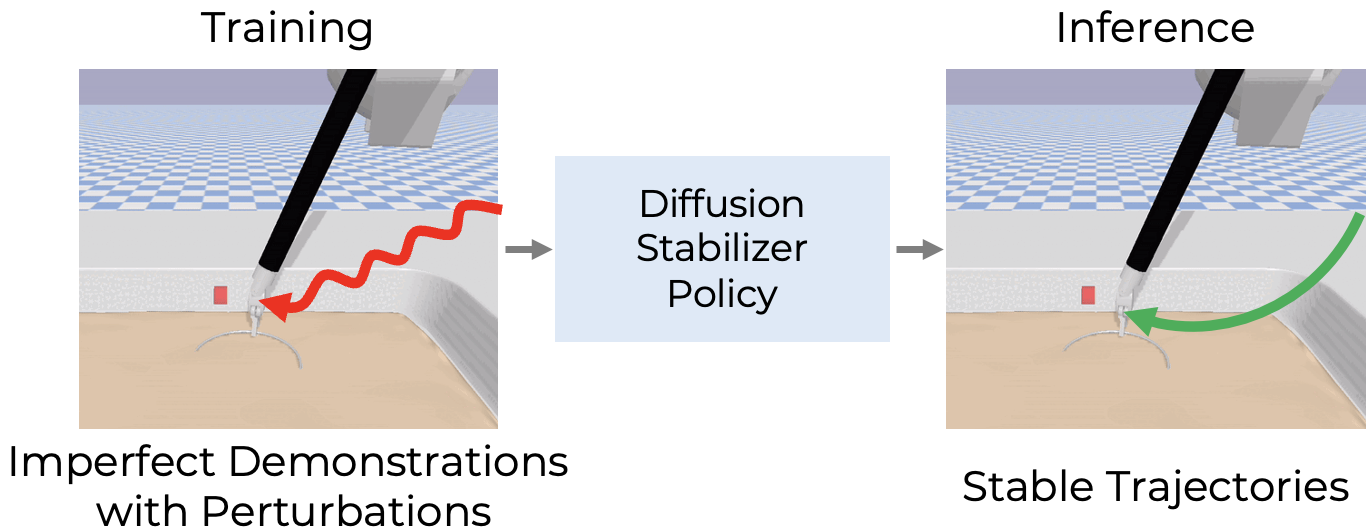}
\caption{\textbf{Overview.} Our diffuion-based policy learning framework learns a diffusion stabilizer which can filter the perturbed data.}
\label{fig:teaser1}
\vspace{-4ex}
\end{figure}

\begin{figure*}[t]
\centering
\includegraphics[width=0.9\textwidth]{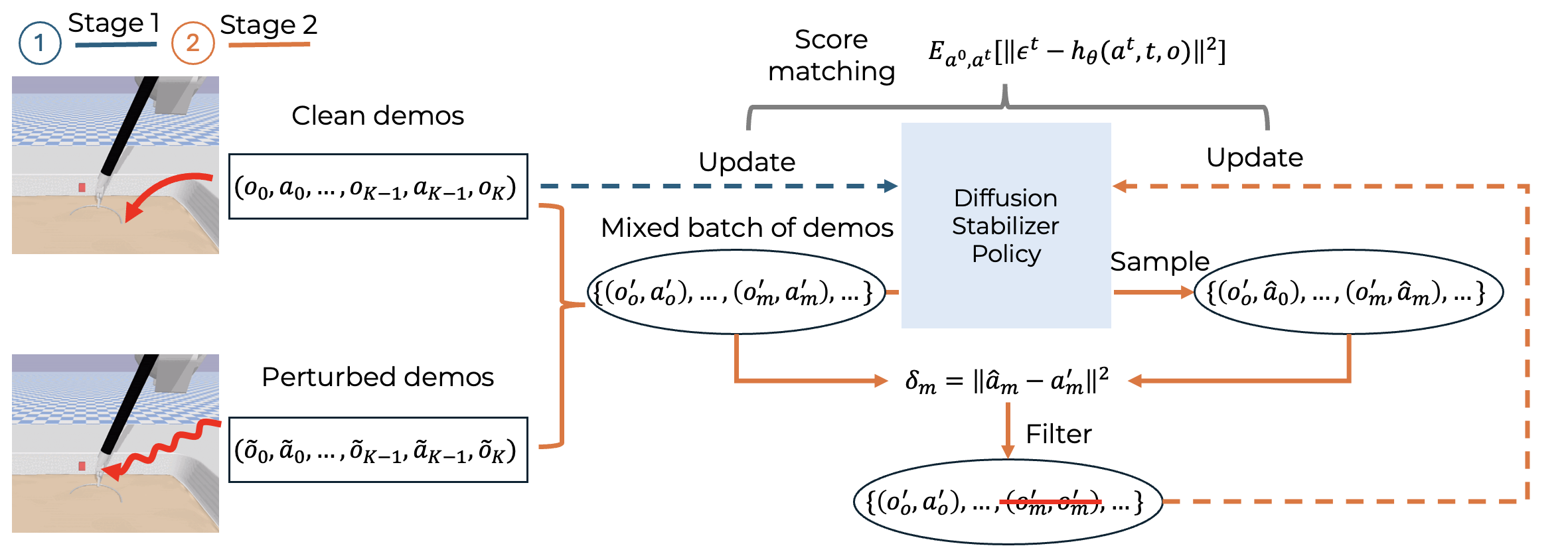}
\caption{The overall training framework of \textbf{Diffusion Stabilizer Policy.} Our diffusion-based policy learning framework first trains a diffusion stabilizer policy with only clean data.
The mixed batch of clean and perturbed data is filtered by the diffusion policy according to the error between predicted actions and the actions from the mixed dataset.
The diffusion policy is continuously updated with the filtered data.}
\label{fig:teaser2}
\vspace{-3ex}
\end{figure*}

Previous work \cite{pmlr-v155-kim21a, li20223dperceptionbasedimitation, MPD, Kawaharazuka_2024, SRT} has explored different imitation learning algorithms with high-quality demonstrations in solving surgical tasks.
Due to the inherent property of mimicking the training dataset using imitation learning, perturbed data could even be detrimental to the performance \cite{SRT}.
To avoid the potential issue caused by perturbed data and fully leverage perturbed data, we propose a diffusion-based policy learning framework, called Diffusion Stabilizer Policy (\ours).
As shown in \Cref{fig:teaser1}, this framework allows filtering the imperfect demonstrations and thus making it possible to train with a combination of clean and perturbed data.
To specify, we consider dealing with different kinds of imperfect demonstrations in which some of the expert actions are perturbed by some noise.
This scenario is common in real-world applications where accidental operation happens or noise of recording device exists during demonstration collection.
The key idea in our framework is to first train the diffusion stabilizer policy with only clean data and then apply it as a filter to process the perturbed data.
By a comprehensive evaluation on different surgical tasks in SurRoL \cite{SurRoL} platform, we demonstrate superior performance of using this diffusion-based policy learning framework when facing both clean data and perturbed data of different types.
The main contributions of our work are summarized:
\begin{itemize}
    \item We propose a diffusion-based policy learning framework for surgical robot manipulations.
    The framework is able to learn stable manipulation even when perturbations are present in the demonstration.
    \item We demonstrate the performance of the proposed framework under two types of perturbation: action-level perturbation and trajectory-level perturbation.
    Our method achieves an overall \textbf{31\%} performance gain on average success rate under action-level perturbations and \textbf{28\%} performance gain under trajectory-level perturbations respectively.
    \item We run real-world experiments to demonstrate that the policy trained with imperfect demonstrations can still solve the tasks.
\end{itemize}

\section{Related Work}
\paragraph{Surgical Simulation Environments}
Surgical robots have earned significant attention from researchers and surgeons.
Due to the inherent challenges and risks associated with directly conducting real-world experiments, simulation platforms have become an essential tool for evaluating and validating robot learning algorithms before real-world deployment.
Different simulation platforms \cite{SurRoL, lapgym, SurgicalAI} have been developed to provide controlled, reproducible environments containing various surgical robotic tasks.
In this paper, we focus specifically on the SurRoL platform \cite{SurRoL}, which offers a combination of task diversity, realism, and alignment with real-world dVRK setups.
Additionally, different baselines \cite{DEX} varying from reinforcement learning to imitation learning have been evaluated in the environments from SurRoL.

\paragraph{Diffusion Policy}
Diffusion model is first proposed as a generative model for image generation \cite{DDPM, score_matching, latent_diffusion}.
By learning the score function of the image data distribution, new images can be generated by first sampling from a normal distribution and then solving a reverse-time SDE \cite{score_matching}.
Recently, the ability of diffusion model to learn a data distribution is exploited in imitation learning for solving robotics tasks \cite{diffusion_policy, 3D_diffusion_policy, rdt-1b}.
A conditional diffusion policy $\pi(a|o)$ is learned from expert demonstrations.
In the domain of surgical robots, the application of diffusion policies in surgical tasks is an active research direction \cite{MPD, SRT, surgical_embodied_ai}.
MPD \cite{MPD} combines diffusion policy and movement primitives \cite{ProDMP} to gain gentle manipulation skills.
SRT \cite{SRT} focuses on solving the issue of imprecise joint measurements in real-world setup by introducing a relative action formulation.
A comprehensive evaluation of robot learning methods on different surgical tasks are presented in \cite{surgical_embodied_ai}.

\paragraph{Diffusion Training with Noisy Data}
The performance of imitation learning highly relies on the quality of the dataset.
A Gaussian policy, commonly exploited in reinforcement learning \cite{sac}, can resist the noise in the dataset by learning the "mean" of the noisy action.
In contrast, the ability of a diffusion model to learn multi-modality in the dataset might hinder its capacity to resist perturbations.
This requires diffusion model to be trained on a high quality dataset to achieve satisfactory performance.
There is research work on learning with corrupted data \cite{daras2023ambient} or noisy label \cite{DBLP:conf/iclr/NaKBLKKM24} in training diffusion for image generation.
Ambient diffusion \cite{daras2023ambient} requires knowing the corruption matrix in advance for learning with corrupted data, which is often infeasible in the context of diffusion policy.
In Label-Noise Robust Diffusion Models \cite{DBLP:conf/iclr/NaKBLKKM24}, the noise in label corresponds to the noise in observations (conditions) to the diffusion policy, which is not compatible with our setting of having noise in the actions.
Overall, learning from noisy data remains under-explored, particularly for diffusion policies in surgical environments.
% All in all, it is under-explored that learning with noisy data especially for diffusion policy in surgical environments. 

\section{Method}
\label{sec:method}

To counteract the potential issue of training with perturbed data,
we propose a new framework which allows training a diffusion model with a combination of clean and perturbed data, as shown in \Cref{fig:teaser2}.
To specify, this framework trains an diffusion stabilizer policy with only clean data in the first stage.
This diffusion stabilizer policy then continues to be updated with a mixture of original clean data and different kinds of perturbed data, which is filtered by the stabilizer policy.

\paragraph{Problem Setting}
A clean dataset $D=\{\tau_i=(o_{0,\ldots,K}, a_{0,\ldots,K-1}), i=1,\ldots,N\}$ with $N$ demonstrations $\tau_i$ of length $K$ and a perturbed dataset $\tilde{D}=\{\tilde{\tau}_i=(\tilde{o}_{0,\ldots,K}, \tilde{a}_{0,\ldots,K-1}), i=1,\ldots,\tilde{N}\}$ with $\tilde{N}$ trajectories $\tilde{\tau}_i$ of the same length are generated in the simulated surgical environments. 
The goal is to learn a policy $\pi(a|o)$ by imitation learning on both clean data $D$ and  perturbed data $\tilde D$.

We designed two categories of perturbation: action-level perturbation and trajectory-level perturbation.
For action-level perturbation, perturbed actions $\tilde{a}$ are generated by adding noise from a variety of distributions, such as the normal distribution $\epsilon \sim \mathcal{N}(\eta, \sigma)$ to a fixed number of expert/optimal actions $a^*$: $\tilde{a} = a^*+\epsilon$ within a trajectory.
We considered noise distributions including Gaussian, poisson, and uniform noise. 
This kind of noise corresponds to the noise from the device used for recording the data when collecting demonstrations in a real-world scenario.

Trajectory-level perturbation is used to mimic the situation in which surgeons fail and retry when collecting the demonstrations.
For example, in the needle-picking scenario, a perturbed demonstration at the trajectory level may include a trajectory where the surgical robot initially approaches the needle badly, retracts, and reattempts successfully.
This type of trajectories departs from the optimal path while still achieving the task.
We have defined a set of these trajectory-level perturbation for 6 surgical tasks from the SurRoL environments, including both PSM and Bi-PSM setups, with detailed descriptions provided in the appendix. 

\paragraph{Overall framework}
Imperfect data with perturbations might be detrimental to the performance of the trained diffusion models especially when using limited demonstrations.
The main idea of {\ours} is to first train a diffusion stabilizer policy only on clean dataset and continue to train it with the clean data and the perturbed data after filtering.
In the first stage, we train a diffusion policy $\pi_\theta(a|o)$ conditioned on the observations, by learning the conditional score function of the data distribution.
The score function is approximated by a diffusion model $h$ parameterized by a Multi-Layer Perceptron (MLP) with parameters $\theta$.

The training of this diffusion policy contains a diffusion process and a denoising process.
In the diffusion process, scheduled Gaussian noise with variance $\beta^t$ is gradually added to the clean action $a^0$ at diffusion step $t$:
\begin{equation}
    q(a^t|a^{t-1}) = \mathcal{N}(a^t;\sqrt{1-\beta^t}a^{t-1},\beta^t\mathbf{I}).
\end{equation}
To avoid confusion, we use the superscript $t$ to indicate the diffusion step, which is different from the subscript $k$ indicating the time step in a trajectory.
With the noisy actions $a^t$, the diffusion model is trained to predict the noise added to it given the diffusion step t and the observation $o$. 
The following loss function is used to train the diffusion model:
\begin{equation}
\mathcal{L} = \mathbb{E}_{o, a\sim D} \Big[ \mathbb{E}_{a^0, a^t}||\epsilon^t-h_\theta(a^t,t,o)||^2 \Big],
\end{equation}
where $(o,a)$ are transitions sampled from the clean dataset $D$, $a^t$ is the noisy action and $\epsilon^t$ is the noise added at diffusion step $t$.

To sample action from the diffusion policy $\pi_\theta(a|o)$, we need to first sample from a Gaussian distribution to get a noisy action $a^t$, and then repeat the denoising step with the learned score function $h_\theta$:
\begin{equation}
    a^{t-1} = \alpha_1 ( a^t - \alpha_2 h_\theta(a^t, t, o) ) + \mathcal{N}(0, \alpha_3 \mathbf{I}),
\end{equation}
where $\alpha_1$, $\alpha_2$, $\alpha_3$ are all constant related to the noise scheduler, only depending on $t$, used in the diffusion process.

Given that the trained diffusion model already encapsulates knowledge about the underlying data distribution after first stage, it can be utilized to detect abnormal data by calculating the error between the actions predicted by it and those in the perturbed dataset.
To filter the perturbed data within a training batch $B=\{(o'_m,a'_m), m=1,\ldots,M\}$ of M samples from the mixed dataset $D'=D+\tilde{D}$, the trained diffusion model is applied to predict an action $\hat{a}_m$ given each observation $o'_m$:
\begin{equation}
    \hat{a}_m \sim \pi_\theta (\cdot|o'_m), B=\{(o'_m,a'_m), m=1,\ldots,M\} \sim D'   
\end{equation}
The error $\delta_m$ and a threshold $\gamma$ are used to filter the state action pair $(o'_m, a'_m)$ in the noisy data set.
\begin{equation}
\begin{aligned}
    \delta_m &= ||\hat{a}_m-a'_m||^2 
    % \hat{\mu} = \frac{1}{K} \sum_{k=0}^{K-1} \delta_k, & \hat{\sigma}^2 = \frac{1}{K-1} \sum_{k=1}^{K-1} (\delta_k - \hat{\mu})^2\\
    % \gamma &= \hat{\mu}
\label{eq:threshold}
\end{aligned}
\end{equation}
If the error $\delta_m$ is larger than the threshold $\gamma$, the transition will not be used for calculating the loss for this batch.
The process of sampling data from the mixed dataset, filtering data with current diffusion stabilizer policy and updating it with the filtered data is repeated until the end of training.

\paragraph{Implementation Details}
We implement our diffusion policy based on CleanDiffuser \cite{dong2024cleandiffuser}, and adopt the default hyperparameter settings.
A four-layer MLP with hidden dimension of 512 is used as the diffusion model.
Actions and observations are processed by two different two-layer MLPs respectively into action embeddings and observation conditions with the same dimension of 128.
Variance Preserving (VP) continuous-time Stochastic Differential Equation (SDE) \cite{score_matching} is used in the forward process.
And the reversed process is solved by a discretized reverse-time VP SDE \cite{DDPM} with 5 denoising steps.
The diffusion model is optimized by AdamW \cite{AdamW} with a learning rate of $2*10^{-4}$.
If not specified, the diffusion policy is trained for 100,000 gradient steps with batch size of 256.
We test our method on a computational node with a RTX 4090 GPU and an AMD EPYC 7763 CPU. Specifically, training the diffusion stabilizer policy for 50,000 steps in the first stage requires approximately 30 to 40 minutes. The second stage, which involves online filtering and continuous updating for another 50,000 steps, takes an additional 30 to 40 minutes. This duration varies depending on task complexity—specifically, the dimensionality of the observation and action spaces.
% It takes about one hour to train a single policy for one surgical task.

\section{Experiments}
\begin{table*}[ht]
\caption{Performance of different learning algorithms in surgical environments}
\vspace{-5ex}
\label{Tab:comparing with baselines}
\begin{center}
\resizebox{\textwidth}{!}{
\begin{tabular}{@{}cllllllllllll@{}}
\toprule
\multicolumn{2}{c}{}                                                                & \multicolumn{2}{c}{Reinforcement Learning}                                                                                     & \multicolumn{3}{c}{Imitation Learning}                                                                                                                                                           & \multicolumn{5}{c}{Demonstration-guided RL}                                                                                                                                                                                                                                                                                        &                                                             \\ \midrule
\multicolumn{2}{c}{Task}                                                            & SAC\cite{sac}                                                           & DDPG\cite{ddpg}                                                           & BC\cite{bc}                                                             & SQIL\cite{sqil}                                                           & VINN \cite{vinn}                                                          & DDPGBC \cite{ddpgher}                                                        & AMP\cite{amp}                                                           & CoL\cite{col}                                                            & AWAC \cite{awac}                                                          & DEX\cite{DEX}                                                            & \textbf{{\ours}(proposed)}                                                   \\ \midrule
                         & \cellcolor[HTML]{EFEFEF}Aggregate                        & \cellcolor[HTML]{EFEFEF}0.99\scriptsize{($\pm$.03)}                        & \cellcolor[HTML]{EFEFEF}0.99\scriptsize{($\pm$.02)}                        & \cellcolor[HTML]{EFEFEF}\textbf{1.00}\scriptsize{($\pm$.00)}               & \cellcolor[HTML]{EFEFEF}0.24\scriptsize{($\pm$.00)}                        & \cellcolor[HTML]{EFEFEF}0.58\scriptsize{($\pm$.06)}                        & \cellcolor[HTML]{EFEFEF}\textbf{1.00}\scriptsize{($\pm$.00)}                        & \cellcolor[HTML]{EFEFEF}\textbf{1.00}\scriptsize{($\pm$.01)}                        & \cellcolor[HTML]{EFEFEF}\textbf{1.00}\scriptsize{($\pm$.00)}                        & \cellcolor[HTML]{EFEFEF}0.99\scriptsize{($\pm$.01)}                        & \cellcolor[HTML]{EFEFEF}\textbf{1.00}\scriptsize{($\pm$.00)}                        & \cellcolor[HTML]{EFEFEF}0.97\ci{.01}
\\
                         & ECMReach                                                                 & \textbf{1.00}\scriptsize{($\pm$.06)}                                       & \textbf{1.00}\scriptsize{($\pm$.00)}                                       & \textbf{1.00}\scriptsize{($\pm$.00)}                                       & 0.07\scriptsize{($\pm$.04)}                                                & 0.49\scriptsize{($\pm$.10)}                                                & \textbf{1.00}\scriptsize{($\pm$.00)}                                       & 0.99\scriptsize{($\pm$.02)}                                                & \textbf{1.00}\scriptsize{($\pm$.00)}                                       & \textbf{1.00}\scriptsize{($\pm$.00)}                                       & \textbf{1.00}\scriptsize{($\pm$.00)}                                       &
                         0.97\ci{.02}
\\
                         & StaticTrack                                                              & 
                         0.92\scriptsize{($\pm$.14)}                                                & 0.98\scriptsize{($\pm$.05)}                                                & \textbf{1.00}\scriptsize{($\pm$.00)}                                       & 0.43\scriptsize{($\pm$.26)}                                                & 0.56\scriptsize{($\pm$.10)}                                                & \textbf{1.00}\scriptsize{($\pm$.00)}                                       & 0.97\scriptsize{($\pm$.03)}                                                & \textbf{1.00}\scriptsize{($\pm$.00)}                                       & \textbf{1.00}\scriptsize{($\pm$.00)}                                       & \textbf{1.00}\scriptsize{($\pm$.00)}                                       & 
                         0.94\ci{.01}
\\
\multirow{-4}{*}{\rotatebox{90}{ECM}}                                                               
                        & MisOrient                                                                 & \textbf{1.00}\scriptsize{($\pm$.00)}                                        & \textbf{1.00}\scriptsize{($\pm$.00)}                                        & \textbf{1.00}\scriptsize{($\pm$.00)}                                        & 0.56\scriptsize{($\pm$.10)}                                                 & 0.50\scriptsize{($\pm$.11)}                                                 & 0.99\scriptsize{($\pm$.02)}                                                 & 0.98\scriptsize{($\pm$.02)}                                                 & 0.99\scriptsize{($\pm$.02)}                                                 & 0.98\scriptsize{($\pm$.03)}                                                 & 0.99\scriptsize{($\pm$.02)}                                                 & 
                        \textbf{1.00}\ci{.01}
\\ \midrule
                         & \cellcolor[HTML]{EFEFEF}Aggregate                        & \cellcolor[HTML]{EFEFEF}0.00\scriptsize{($\pm$.00)}                        & \cellcolor[HTML]{EFEFEF}0.00\scriptsize{($\pm$.00)}                        & \cellcolor[HTML]{EFEFEF}0.40\scriptsize{($\pm$.05)}                        & \cellcolor[HTML]{EFEFEF}0.00\scriptsize{($\pm$.00)}                        & \cellcolor[HTML]{EFEFEF}0.02\scriptsize{($\pm$.02)}                        & \cellcolor[HTML]{EFEFEF}0.80\scriptsize{($\pm$.04)}                        & \cellcolor[HTML]{EFEFEF}0.00\scriptsize{($\pm$.00)}                        & \cellcolor[HTML]{EFEFEF}0.85\scriptsize{($\pm$.06)}                        & \cellcolor[HTML]{EFEFEF}0.46\scriptsize{($\pm$.19)}                        & \cellcolor[HTML]{EFEFEF}0.89\scriptsize{($\pm$.03)}                        & \cellcolor[HTML]{EFEFEF}\textbf{0.98\ci{.01}}\\
                         & NeedleReach                                              & \textbf{1.00}\scriptsize{($\pm$.00)}                                                & \textbf{1.00}\scriptsize{($\pm$.00)}                                                & \textbf{1.00}\scriptsize{($\pm$.00)}                                                & 0.07\scriptsize{($\pm$.09)}                                                & 0.89\scriptsize{($\pm$.06)}                                                & \textbf{1.00}\scriptsize{($\pm$.00)}                                                & 0.99\scriptsize{($\pm$.02)}                                                & \textbf{1.00}\scriptsize{($\pm$.00)}                                               & 0.94\scriptsize{($\pm$.20)}                                                & \textbf{1.00}\scriptsize{($\pm$.00)}                                                & \textbf{1.00}\ci{.01}
\\
                         & GauzeRetrieve                                            & 0.00\scriptsize{($\pm$.00)}                                                & 0.00\scriptsize{($\pm$.00)}                                                & 0.07\scriptsize{($\pm$.05)}                                                & 0.00\scriptsize{($\pm$.00)}                                                & 0.01\scriptsize{($\pm$.02)}                                                & 0.63\scriptsize{($\pm$.11)}                                                & 0.00\scriptsize{($\pm$.00)}                                                & 0.71\scriptsize{($\pm$.16)}                                                & 0.43\scriptsize{($\pm$.43)}                                                & \textbf{0.73}\scriptsize{($\pm$.12)}                                                & \textbf{0.73}\ci{.08}
\\
                         & NeedlePick                                               & 0.00\scriptsize{($\pm$.00)}                                                & 0.00\scriptsize{($\pm$.00)}                                                & 0.21\scriptsize{($\pm$.06)}                                                & 0.00\scriptsize{($\pm$.00)}                                                & 0.02\scriptsize{($\pm$.02)}                                                & 0.91\scriptsize{($\pm$.05)}                                                & 0.00\scriptsize{($\pm$.00)}                                                & 0.96\scriptsize{($\pm$.05)}                                                & 0.26\scriptsize{($\pm$.33)}                                                & 0.94\scriptsize{($\pm$.05)}                                                & \textbf{0.99\ci{.02}}
\\
\multirow{-5}{*}{\rotatebox{90}{PSM}}    & PegTransfer                                              & 0.00\scriptsize{($\pm$.00)}                                                & 0.00\scriptsize{($\pm$.00)}                                                & 0.56\scriptsize{($\pm$.11)}                                                & 0.02\scriptsize{($\pm$.05)}                                                & 0.05\scriptsize{($\pm$.04)}                                                & 0.48\scriptsize{($\pm$.22)}                                                & 0.00\scriptsize{($\pm$.00)}                                                & 0.58\scriptsize{($\pm$.23)}                                                & 0.31\scriptsize{($\pm$.32)}                                                & 0.73\scriptsize{($\pm$.20)}                                                & \textbf{0.99\ci{.02}}
\\ \midrule
                         & \cellcolor[HTML]{EFEFEF}{\color[HTML]{333333} Aggregate} & \cellcolor[HTML]{EFEFEF}{\color[HTML]{333333} 0.00\scriptsize{($\pm$.00)}} & \cellcolor[HTML]{EFEFEF}{\color[HTML]{333333} 0.00\scriptsize{($\pm$.00)}} & \cellcolor[HTML]{EFEFEF}{\color[HTML]{333333} 0.08\scriptsize{($\pm$.04)}} & \cellcolor[HTML]{EFEFEF}{\color[HTML]{333333} 0.00\scriptsize{($\pm$.00)}} & \cellcolor[HTML]{EFEFEF}{\color[HTML]{333333} 0.00\scriptsize{($\pm$.00)}} & \cellcolor[HTML]{EFEFEF}{\color[HTML]{333333} 0.00\scriptsize{($\pm$.00)}} & \cellcolor[HTML]{EFEFEF}{\color[HTML]{333333} 0.00\scriptsize{($\pm$.00)}} & \cellcolor[HTML]{EFEFEF}{\color[HTML]{333333} 0.00\scriptsize{($\pm$.00)}} & \cellcolor[HTML]{EFEFEF}{\color[HTML]{333333} 0.00\scriptsize{($\pm$.00)}} & \cellcolor[HTML]{EFEFEF}{\color[HTML]{333333} 0.39\scriptsize{($\pm$.11)}} & \cellcolor[HTML]{EFEFEF}\textbf{0.83}\ci{.05}\\
                         & NeedleRegrasp                                            & 0.00\scriptsize{($\pm$.00)}                                                & 0.00\scriptsize{($\pm$.00)}                                                & 0.09\scriptsize{($\pm$.03)}                                                & 0.01\scriptsize{($\pm$.00)}                                                & 0.01\scriptsize{($\pm$.02)}                                                & 0.05\scriptsize{($\pm$.08)}                                                & 0.00\scriptsize{($\pm$.00)}                                                & 0.04\scriptsize{($\pm$.07)}                                                & 0.00\scriptsize{($\pm$.00)}                                               & 0.63\scriptsize{($\pm$.19)}                                                & \textbf{0.84}\ci{.07}
\\
\multirow{-3}{*}{\rotatebox{90}{Bi-PSM}} & BiPegTransfer                                            & 0.00\scriptsize{($\pm$.00)}                                                & 0.00\scriptsize{($\pm$.00)}                                                & 0.09\scriptsize{($\pm$.05)}                                                & 0.00\scriptsize{($\pm$.00)}                                                & 0.00\scriptsize{($\pm$.00)}                                                & 0.00\scriptsize{($\pm$.00)}                                                & 0.00\scriptsize{($\pm$.00)}                                                & 0.01\scriptsize{($\pm$.02)}                                                & 0.00\scriptsize{($\pm$.00)}                                                & 0.18\scriptsize{($\pm$.14)}                                                & \textbf{0.82}\ci{.08}
\\ \midrule
\multicolumn{1}{l}{}     & \cellcolor[HTML]{EFEFEF}Overall                          & \cellcolor[HTML]{EFEFEF}0.46\scriptsize{($\pm$.03)}                        & \cellcolor[HTML]{EFEFEF}0.45\scriptsize{($\pm$.01)}                        & \cellcolor[HTML]{EFEFEF}0.68\scriptsize{($\pm$.02)}                        & \cellcolor[HTML]{EFEFEF}0.02\scriptsize{($\pm$.02)}                        & \cellcolor[HTML]{EFEFEF}0.24\scriptsize{($\pm$.03)}                        & \cellcolor[HTML]{EFEFEF}0.83\scriptsize{($\pm$.05)}                        & \cellcolor[HTML]{EFEFEF}0.48\scriptsize{($\pm$.01)}                        & \cellcolor[HTML]{EFEFEF}0.87\scriptsize{($\pm$.03)}                        & \cellcolor[HTML]{EFEFEF}0.58\scriptsize{($\pm$.08)}                        & \cellcolor[HTML]{EFEFEF}0.92\scriptsize{($\pm$.02)}                        & \cellcolor[HTML]{EFEFEF}\textbf{0.96\ci{.01}}
\\ \bottomrule
\\
\end{tabular}
}
\end{center}
\vspace{-5ex}
\end{table*}

\begin{table*}[ht]
\centering
\caption{Evaluation result of different perturbation for different tasks}
\vspace{-1ex}
\label{Tab:noise_imperfect}
\resizebox{1\linewidth}{!}{
\begin{tabular}{@{}ll*{12}{c}@{}}
\toprule
 &   \multicolumn{6}{c}{Action-level noise} & \multicolumn{6}{c}{Trajectory-level noise} \\ 
\cmidrule(lr){3-8} \cmidrule(lr){9-14}
 &  &   \multicolumn{2}{c}{Gaussian} & \multicolumn{2}{c}{Poisson} & \multicolumn{2}{c}{Uniform} & \multicolumn{2}{c}{Type 1} & \multicolumn{2}{c}{Type 2} & \multicolumn{2}{c}{Type 3} \\ 
\cmidrule(lr){3-4} \cmidrule(lr){5-6} \cmidrule(lr){7-8} \cmidrule(lr){9-10} \cmidrule(lr){11-12} \cmidrule(lr){13-14}
Category & Task & Perturbed & ours & Perturbed & ours & Perturbed & ours & Perturbed & ours & Perturbed & ours & Perturbed & ours \\ 
\midrule
\multirow{4}{*}{PSM} 
 & NeedleReach    & 0.96\ci{.01} & \ourcell \textbf{1.00\ci{.01}} & 0.23\ci{.04} & \ourcell \textbf{0.98\ci{.02}} & 0.61\ci{.04} & \ourcell \textbf{1.0\ci{.00}} & 0.61\ci{.07} & \ourcell \textbf{1.00\ci{.00}} & \textbf{0.99\ci{.02}} & \ourcell \textbf{0.99\ci{.01}} & -- & \ourcell -- \\
 & NeedlePick     & 0.38\ci{.15} & \ourcell \textbf{0.81\ci{.04}} & 0.69\ci{.06} & \ourcell \textbf{0.83\ci{.12}} & \textbf{0.74\ci{.09}} & \ourcell \textbf{0.74\ci{.08}} & 0.58\ci{.10} & \ourcell \textbf{0.79\ci{.06}} & \textbf{0.49\ci{.11}} & \ourcell 0.43\ci{.11} & 0.19\ci{.05} & \ourcell \textbf{0.60\ci{.04}} \\
 & GauzeRetrieve  & 0.67\ci{.07} & \ourcell \textbf{0.80\ci{.04}} & 0.30\ci{.07} & \ourcell \textbf{0.59\ci{.02}} & 0.71\ci{.05} & \ourcell \textbf{0.85\ci{.07}} & \textbf{0.65\ci{.07}} & \ourcell 0.39\ci{.07} & 0.26\ci{.04} & \ourcell \textbf{0.52\ci{.05}} & 0.05\ci{.02} & \ourcell \textbf{0.18\ci{.04}} \\
 & PegTransfer    & \textbf{0.83\ci{.03}} & \ourcell 0.79\ci{.06} & 0.86\ci{.01} & \ourcell \textbf{0.93\ci{.02}} & 0.89\ci{.02} & \textbf{\ourcell 0.95\ci{.05}} & 0.69\ci{.06} & \ourcell \textbf{0.85\ci{.02}} & \textbf{0.83\ci{.10}} & \ourcell 0.61\ci{.09} & 0.41\ci{.08} & \ourcell \textbf{0.53\ci{.10}} \\
\midrule
\multirow{2}{*}{Bi-PSM}
 & BiPegTransfer  & 0.59\ci{.06} & \ourcell \textbf{0.74\ci{.03}} & 0.49\ci{.05} & \ourcell \textbf{0.73\ci{.09}} & 0.67\ci{.08} & \ourcell \textbf{0.71\ci{.09}} & 0.48\ci{.12} & \ourcell \textbf{0.57\ci{.11}} & 0.49\ci{.08} & \ourcell \textbf{0.57\ci{.07}} & 0.19\ci{.06} & \ourcell \textbf{0.70\ci{.06}} \\
 & NeedleRegrasp  & 0.40\ci{.06} & \ourcell \textbf{0.69\ci{.07}} & 0.35\ci{.11} & \ourcell \textbf{0.60\ci{.04}} & 0.62\ci{.07} & \ourcell \textbf{0.63\ci{.06}} & 0.76\ci{.06} & \ourcell \textbf{0.85\ci{.07}} & 0.63\ci{.06} & \ourcell \textbf{0.72\ci{.05}} & 0.43\ci{.10} & \ourcell \textbf{0.66\ci{.06}} \\
% \midrule
% Overall & & 0.00\ci{0.00} & \ourcell 0.00\ci{0.00} & 0.00\ci{0.00} & \ourcell 0.00\ci{0.00} & 0.00\ci{0.00} & \ourcell 0.00\ci{0.00} & 0.00\ci{0.00} & \ourcell 0.00\ci{0.00} & 0.00\ci{0.00} & \ourcell 0.00\ci{0.00} & 0.00\ci{0.00} & \ourcell 0.00\ci{0.00} \\
\bottomrule
\end{tabular}
}
\vspace{-3ex}
\end{table*}

In this section, we first validate our proposed method (\ours) in different environments of SurRoL \cite{SurRoL} with only clean dataset.
Then we show the performance of our method when two types of perturbations are included in the training set.
Finally, we present an ablation study of our method under different training settings.

\subsection{Experimental Setup}
SurRoL \cite{SurRoL} is a well-designed platform to simulate the real-world dVRK system.
It contains 10 challenging tasks, which can be classified into 3 categories according to the type of the manipulator:
(1) Single-handed patient-sided manipulator (\textbf{PSM});
(2) Bimanual PSM (\textbf{Bi-PSM});
and (3) Endoscopic camera manipulator (\textbf{ECM}).
The observation space contains the position and orientation of both the end-effector and the object, while Cartesian-space control is applied as actions on the end-effector.
% The observation space contains the position and orientation of both the end-effector and object. 
% And Cartesian-space control is applied as actions on the end-effector.
We use the same evaluation metrics as used in the baselines \cite{DEX}, which calculate the interquartile mean (IQM) and estimate the 95\% confidence interval.

\subsection{Main Results}
We start with training a diffusion model using the same number (100 episodes) of expert demonstrations (length of 50) as the baselines \cite{DEX}.
The success rate evaluated in different environments and the aggregated performance are shown in \Cref{Tab:comparing with baselines}.
% \TODO{Double check the performance in GauzeRetrieve.} solved
Our method can maintain comparable performance in relative simple ECM tasks compared to the strong baseline DEX \cite{DEX}, while outperforming all the other baselines in complex tasks.
Especially for long-horizon tasks such as BiPegTransfer, which requires the bimanual agent to collaborate on transferring the peg, diffusion policy achieves +355\% IQM.
As indicated by the aggregated performance for all tasks, leveraging diffusion policy can really push the success rate closer to $100\%$.

To demonstrate that DSP holds a clear advantage over traditional diffusion-based methods, we use the standard Diffusion Policy \cite{diffusion_policy} as our primary baseline in the perturbation experiments. In the following tables, the columns labeled 'Perturbed' denote the performance of this standard Diffusion Policy trained directly on the mixed datasets without our proposed filtering mechanism.

%% this table is using (25,25), thr=mean for action-level and thr=mean-var(strict) for trajectory-level.

\begin{table}[ht]
\vspace{6mm}
\centering
\caption{Evaluation Result of online and offline mode for different tasks}
\vspace{-0.5mm}
\label{Tab:offline_and_online}
\setlength{\tabcolsep}{5pt}
\renewcommand{\arraystretch}{1.05}
\resizebox{1\linewidth}{!}{
\begin{tabular}{@{}ll*{12}{c}@{}}
\toprule
 &  &   \multicolumn{3}{c}{Action-level noise} & \multicolumn{3}{c}{Trajectory-level noise} \\ 
\cmidrule(lr){3-5} \cmidrule(lr){6-8}
Category & Task & Perturbed & offline & online & Perturbed & offline & online \\ 
\midrule
\multirow{4}{*}{PSM} 
 & NeedleReach    &  0.10\ci{.03} & 0.53\ci{.03} & \ourcell \textbf{0.99\ci{.02}} &  0.61\ci{.07} & 0.65\ci{.06} & \ourcell \textbf{1.00\ci{.00}}  \\
 & NeedlePick     &  0.61\ci{.07} & \textbf{0.93\ci{.04}} & \ourcell \textbf{0.93\ci{.04}} &  0.58\ci{.10} & 0.46\ci{.13} & \ourcell \textbf{0.79\ci{.06}}  \\
 & GauzeRetrieve  &  \textbf{0.73}\ci{.08} & \textbf{0.73}\ci{.08} & \ourcell \textbf{0.73\ci{.08}} & \textbf{0.65\ci{.07}} & 0.59\ci{.06} & \ourcell 0.39\ci{.07}  \\
 & PegTransfer    &  \textbf{0.89\ci{.04}} & 0.86\ci{.07} & \ourcell 0.88\ci{.04} & \textbf{0.83\ci{.10}} & 0.81\ci{.05} & \ourcell 0.61\ci{.09}  \\
\midrule
\multirow{2}{*}{Bi-PSM}
 & BiPegTransfer  & 0.52\ci{.04} & 0.50\ci{.09} & \ourcell \textbf{0.57\ci{.03}} &  0.48\ci{.12} & \textbf{0.77\ci{.07}} & \ourcell 0.57\ci{.11}  \\
 & NeedleRegrasp  & 0.49\ci{.04} & \textbf{0.69\ci{.05}} & \ourcell 0.54\ci{.08 }&  0.76\ci{.06} & \textbf{0.88\ci{.05}} & \ourcell 0.85\ci{.07}  \\
\midrule
Aggregate &  & {0.58}\ci{.03} & {0.71}\ci{.03} &  \ourcell\textbf{0.80}\ci{.02} & 0.65\ci{0.03} & \textbf{0.71\ci{0.03}} & 0.69\ci{0.03} \\
\bottomrule
\end{tabular}
}
\vspace{-3ex}
\end{table}

\begin{table}[ht]
\vspace{6mm}
\centering
\caption{Effects of different demonstration amounts}
\label{Tab:different_num_of_demos}
\vspace{-0.5mm}
\setlength{\tabcolsep}{5pt}
\renewcommand{\arraystretch}{1.05}
\resizebox{1\linewidth}{!}
{ 
\begin{tabular}{lccccc}
\toprule\noalign{\vskip -0.2ex}
\multirow{2}{*}{Method}  & \multicolumn{5}{c}{Number of episodes in demonstrations}  \\ \noalign{\vskip -0.7ex}\cmidrule(lr){2-6}\noalign{\vskip -0.7ex}
& 10 epi.  & 25 epi.  & 50 epi.  & 75 epi. & 100 epi.\\ 
    
\noalign{\vskip -0.5ex}\midrule
BC~\cite{bc} & 0.00\ci{.00} & 0.05\ci{.02} & 0.15\ci{.03} & 0.19\ci{.03} & 0.22\ci{.03} \\
SQIL~\cite{sqil} & 0.00\ci{.00} & 0.00\ci{.00} & 0.00\ci{.00} & 0.00\ci{.00} & 0.00\ci{.00} \\
VINN~\cite{vinn} & 0.03\ci{.03} & 0.02\ci{.03} & 0.02\ci{.02} & 0.01\ci{.02} & 0.01\ci{.01} \\
\cdashline{1-6}\noalign{\vskip 0.5ex}
DDPGBC~\cite{ddpgher} & 0.00\ci{.03} & 0.20\ci{.11} & 0.35\ci{.05} & 0.47\ci{.10} & 0.45\ci{.08} \\
AMP~\cite{amp} & 0.00\ci{.00} & 0.00\ci{.00} & 0.00\ci{.00} & 0.00\ci{.00} & 0.00\ci{.00} \\
CoL~\cite{col} & 0.18\ci{.07} & 0.47\ci{.09} & 0.46\ci{.09} & 0.49\ci{.05} & 0.43\ci{.10} \\
AWAC~\cite{awac} & 0.00\ci{.05} & 0.05\ci{.14} & 0.07\ci{.18} & 0.40\ci{.15} & 0.39\ci{.13} \\
DEX ~\cite{DEX} & \textbf{0.24}\ci{.02} & {0.50}\ci{.09} & {0.68}\ci{.07} & {0.78}\ci{.07} & {0.80}\ci{.06} \\
\noalign{\vskip -0.2ex}\midrule\noalign{\vskip -0.2ex}
\ourcell \textbf{DSP (Proposed)}  & \ourcell {0.15}\ci{.02} & \ourcell \textbf{0.67}\ci{.03} & \ourcell \textbf{0.84}\ci{.02} & \ourcell \textbf{0.91}\ci{.02} & \ourcell \textbf{0.92}\ci{.02}
 \\
\noalign{\vskip -0.2ex}
\bottomrule

% \vspace{-0.8cm}
\end{tabular}
}
\end{table}

\begin{table*}[ht]
\vspace{20mm}
\centering
\caption{Evaluation with different perturb ratio and total episode amount for different tasks}
\vspace{6mm}
\label{Tab:noisy_ratio_comparison}
\resizebox{1\linewidth}{!}{
\begin{tabular}{@{}lllllllllllll@{}}
\toprule
 &   \multicolumn{6}{c}{Action-level Perturb} & \multicolumn{6}{c}{Trajectory-level Perturb} \\ 
\cmidrule(lr){2-7} \cmidrule(lr){8-13}
Env           & \multicolumn{2}{c}{$(N,\tilde{N})=(15,15)$} & \multicolumn{2}{c}{$(N,\tilde{N})=(25,25)$} & \multicolumn{2}{c}{$(N,\tilde{N})=(15,35)$} & \multicolumn{2}{c}{$(N,\tilde{N})=(15,15)$}    & \multicolumn{2}{c}{$(N,\tilde{N})=(25,25)$}  & \multicolumn{2}{c}{$(N,\tilde{N})=(15,35)$}        \\ \midrule
models        & perturbed & \ourcell ours & perturbed & \ourcell ours & perturbed & \ourcell ours & perturbed & \ourcell ours & perturbed & \ourcell ours & perturbed & \ourcell ours \\ \midrule

NeedleReach   &  0.03\ci{0.02} & \ourcell \textbf{0.96\ci{0.02}} & 0.10\ci{0.03}  & \ourcell \textbf{0.99\ci{0.02}} & 0.41\ci{0.12}  & \ourcell \textbf{0.99\ci{0.01}} & 0.43\ci{0.08}  & \ourcell \textbf{0.99\ci{0.02}}  & 0.61\ci{0.07}  & \ourcell \textbf{1.00\ci{0.00}}  & 0.27\ci{0.06}  & \ourcell \textbf{0.94\ci{0.01}}
\\
GauzeRetrieve &  0.40\ci{0.02} & \ourcell \textbf{0.73\ci{0.03}} & \textbf{0.73\ci{0.08}}  & \ourcell \textbf{0.73\ci{0.08}} & 0.51\ci{0.02}  & \ourcell \textbf{0.76\ci{0.07}} & \textbf{0.31\ci{0.07}}  & \ourcell 0.21\ci{0.04}  & \textbf{0.65\ci{0.07}}  & \ourcell 0.39\ci{0.07}  & \textbf{0.23\ci{0.04}}  & \ourcell 0.13\ci{0.03}
\\
NeedlePick    &  0.06\ci{0.07} & \ourcell \textbf{0.55\ci{0.04}} & 0.61\ci{0.07}  & \ourcell \textbf{0.93\ci{0.04}} & 0.55\ci{0.09}  & \ourcell \textbf{0.96\ci{0.02}} & 0.46\ci{0.13}  & \ourcell \textbf{0.63\ci{0.08}}  & 0.58\ci{0.10}  & \ourcell \textbf{0.79\ci{0.06}}  & 0.48\ci{0.08}  & \ourcell \textbf{0.55\ci{0.11}}
\\
PegTransfer   &  \textbf{0.45\ci{0.08}} & \ourcell 0.37\ci{0.03} & \textbf{0.89\ci{0.04}}  & \ourcell 0.88\ci{0.04} & \textbf{0.87\ci{0.04}}  & \ourcell 0.81\ci{0.06} & \textbf{0.63\ci{0.10}}  & \ourcell 0.36\ci{0.11}  & \textbf{0.83\ci{0.10}}  & \ourcell 0.61\ci{0.09}  & \textbf{0.77\ci{0.05}}  & \ourcell 0.37\ci{0.07}
\\ \midrule
NeedleRegrasp &  0.39\ci{0.05} & \ourcell \textbf{0.67\ci{0.04}} & 0.49\ci{0.04}  & \ourcell \textbf{0.54\ci{0.08}} & 0.27\ci{0.04}  & \ourcell \textbf{0.35\ci{0.02}} & 0.49\ci{0.05}  & \ourcell \textbf{0.51\ci{0.06}}  & 0.76\ci{0.06}  & \ourcell \textbf{0.85\ci{0.07}}  & \textbf{0.71\ci{0.04}}  & \ourcell 0.61\ci{0.08} 
\\
BiPegTransfer &  0.41\ci{0.10} & \ourcell \textbf{0.64\ci{0.09}} & 0.52\ci{0.04}  & \ourcell \textbf{0.57\ci{0.03}} & 0.59\ci{0.02}  & \ourcell \textbf{0.75\ci{0.07}} & 0.53\ci{0.07}  & \ourcell \textbf{0.54\ci{0.08}}  & 0.48\ci{0.12}  & \ourcell \textbf{0.57\ci{0.11}}  & \textbf{0.73\ci{0.09}}  & \ourcell 0.43\ci{0.07}
\\\midrule
Aggregate    &  0.32\ci{0.03} & \ourcell \textbf{0.65\ci{0.02}} & 0.58\ci{0.03}  & \ourcell \textbf{0.80\ci{0.02}} & 0.52\ci{0.03}  & \ourcell \textbf{0.81\ci{0.03}} & 0.48\ci{0.03}  & \ourcell \textbf{0.51\ci{0.03}}  & 0.65\ci{0.03}  & \ourcell \textbf{0.69\ci{0.03}}  & \textbf{0.56\ci{0.03}}  & \ourcell 0.49\ci{0.04}    
\\ \bottomrule
\end{tabular}
}
% \vspace{-3ex}
\end{table*}

%%%%%%%%%%%%%% old table for different noise type %%%%%%%%%%%%%%
% \begin{table*}[ht]
% \centering
% \caption{Training result of different perturbation for different tasks}
% \vspace{-1ex}
% \label{Tab:noise_imperfect}
% \resizebox{1\linewidth}{!}{
% \begin{tabular}{@{}lllllllllll@{}}
% \toprule
%  &  &   \multicolumn{4}{c}{perturbation noise} & \multicolumn{4}{c}{imperfect demonstration} \\ \midrule
% Category & Task & stage1 & Gaussian & Poisson & Uniform & imperfect1 & imperfect2 & imperfect3 &  \\ \midrule
% \multirow{4}{*}{PSM} 
%  & NeedleReach  & 0.99\ci{.01} & 0.99\ci{.01} & 0.86\ci{.07} & 0.95\ci{.02} & 0.59\ci{.03} & 0.77\ci{.06} & -- \\
%  & NeedlePick   & 0.87\ci{.09} & 0.81\ci{.03} & 0.83\ci{.12} & 0.74\ci{.08} & 0.42\ci{.09} & 0.37\ci{.15} & 0.23\ci{.04}\\
%  & GauzeRetrieve & 0.24\ci{.05} & 0.80\ci{.04} & 0.59\ci{.02} & 0.85\ci{.07} & 0.62\ci{.06} & 0.31\ci{.07} & 0.11\ci{.07} & \\
%  & PegTransfer  &  0.59\ci{.04} & 0.78\ci{.04} & 0.93\ci{.02} & 0.95\ci{.05} & 0.70\ci{.05} & 0.79\ci{.06} & 0.49\ci{.11} \\
% \midrule
% \multirow{2}{*}{Bi-PSM}
%  & BiPegTransfer &  0.26\ci{.04} & 0.41\ci{.05} &0.00\ci{.00} & 0.00\ci{.00} & 0.75\ci{.03} &  &  \\
%  & NeedleRegrasp & 0.84\ci{.03} & 0.69\ci{.06} & 0.60\ci{.04} & 0.63\ci{.06} & 0.73\ci{.06} & &  \\
% \midrule
% Aggregate &  &  &  &  &  &  &  &   &  &  \\
% \bottomrule
% \end{tabular}
% }
% \vspace{-3ex}
% \end{table*}
%%%%%%%%%%%%%%%%%%%%%%%%%%%%%%%%%%%%%%%%%%%%%%%%%%%%%%%%

Building upon this experimental setup, we further evaluate the robustness and generalization of our framework on imperfect demonstration with different types of perturbation.
% Specifically, we conduct experiments under the $(N,\tilde{N})=(25,25)$ mixed data configuration using online filtering mode. 
In the experiments, we consider three action-level perturbation types: gaussian, uniform and poisson noise, as described in the \Cref{sec:method}.
Moreover, we include structured trajectory-level demonstrations that reflect realistic, task-specific suboptimal strategies, such as misapproaches and recovery behaviors.
The detailed descriptions for each type of imperfect demonstrations according to their ID are presented in the Appendix.
For the threshold, we base on the empirical mean $\hat{\mu}$ and the empirical variance $\hat{\sigma}$:
\begin{equation}
\begin{aligned}
    \hat{\mu} = \frac{1}{M} \sum_{m=1}^{M} \delta_m, & \hat{\sigma}^2 = \frac{1}{M-1} \sum_{m=1}^{M} (\delta_m - \hat{\mu})^2,\\
\end{aligned}
\end{equation}

Two thresholds: $\hat{\mu}$ and $\hat{\mu}-\hat{\sigma}$, are used in the training with action-level perturbation and trajectory-level perturbation, respectively.
The results, summarized in \Cref{Tab:noise_imperfect}, indicate that our filtering mechanism consistently improves policy performance across different noise conditions and imperfect demonstration types.
Notably, our method effectively identifies and suppresses both stochastic perturbations and systematic biases, leading to higher task success rates compared to naive inclusion of perturbed data.
% This demonstrates the capability of our diffusion stabilizer to adaptively handle diverse and realistic forms of noise without requiring prior knowledge of the noise distribution.

\subsection{Ablation Study and Analysis}
In this section, we design experiments to answer the following questions:
1) Does our method successfully filter the perturbed data?
2) Does our method work with different numbers of clean and perturbed data?
3) What are the results of using different thresholds for filtering?

\begin{figure}[t]
    \centering
    \begin{minipage}{0.23\textwidth}
        \centering
        \includegraphics[width=\linewidth]{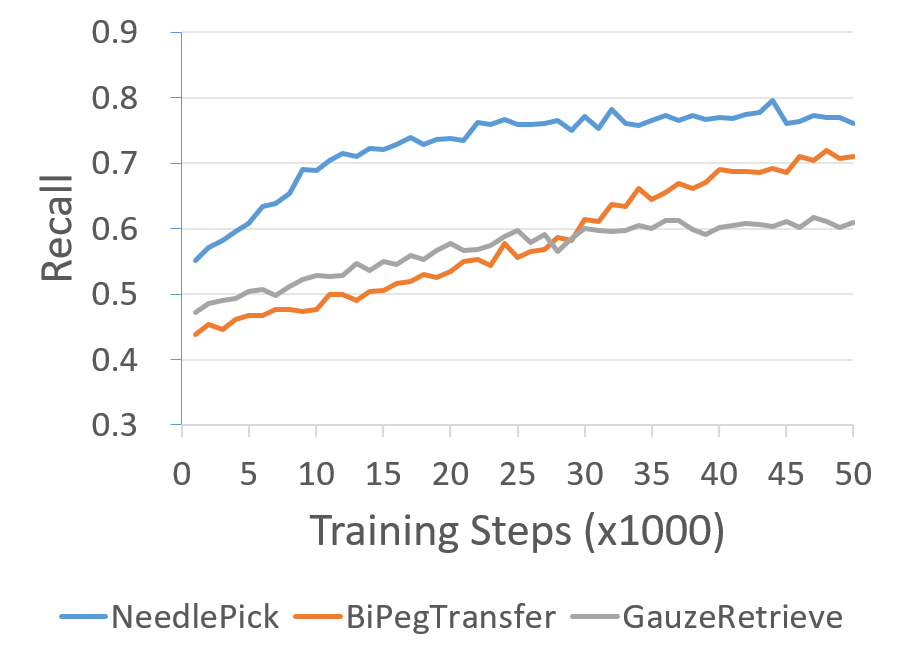}
        % \caption{Recall}
        % \label{fig:ablation_recall}
    \end{minipage}
    % \hfill
    \hspace{1mm}
    \begin{minipage}{0.23\textwidth}
        \centering
        \includegraphics[width=\linewidth]{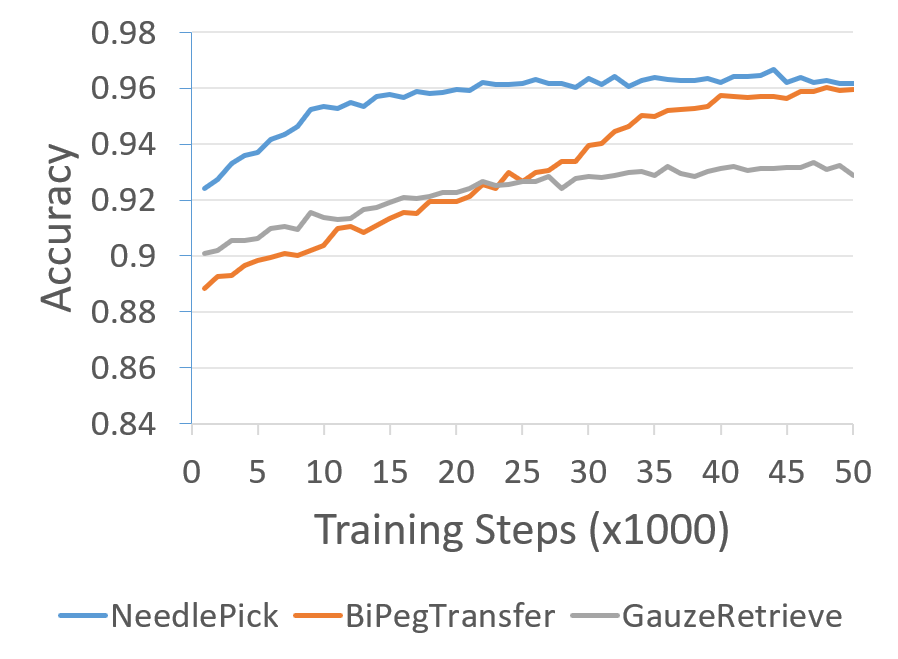}
        % \caption{Accuracy}
        % \label{fig:ablation_accuracy}
    \end{minipage}
    \caption{The perturbed samples generated during data collection and their filtering results are recorded.
    The left figure illustrates the recall of the predictions, representing the percentage of correctly identified perturbed data.
    The right figure depicts the accuracy, indicating the percentage of correctly classified samples across the entire dataset.}
    \label{fig:ablation1}
% \vspace{-3ex}
\end{figure}

\paragraph{Does our method successfully filter the perturbed data?}
We answer the first question by calculating 1) \textbf{the recall:} the percentage of perturbed samples which are correctly filtered by our diffusion stabilizer policy; 2) \textbf{the accuracy:} the percentage of samples which are correctly classified by our diffusion stabilizer policy along the training in the second stage.
The experiments are conducted on a subset of tasks, including 3 representative tasks: NeedlePick, GauzeRetrieve, and BiPegTransfer.
As shown in \Cref{fig:ablation1}, both the recall and accuracy keep increasing along the training and the diffusion stabilizer policy can achieve a high accuracy at the end of training.
Furthermore, two variations of the filtering methods under our framework, marked as "online" and "offline", are also tested. 
In the "offline" mode, the entire perturbed dataset is evaluated and filtered a single time using the frozen model weights obtained at the end of the first stage.
Conversely, the "online" mode continuously evaluates and filters the mixed batch during the second training stage using the actively updating model weights.
This allows dynamically adjusting its filtering criteria as the approximation of the data distribution improves, leading a better classification of borderline samples.
% For "online" mode, as described in the \Cref{sec:method}, the perturbed dataset is filtered by the current diffusion policy during the training.
% In contrast, "offline" mode corresponds to filtering the whole perturbed dataset at once with the model trained after first stage.
The results are shown in \Cref{Tab:offline_and_online}, demonstrating the effectiveness of our method in detecting perturbed data and highlighting the advantages of using an online mode over an offline mode.

% \begin{figure}[t]
%     \centering
%     \begin{minipage}{0.15\textwidth}
%         \centering
%         \includegraphics[width=\linewidth]{figures/NeedlePick_noise2.jpg}
%     \end{minipage}  
%     % \hfill
%     \hspace{1mm}
%     \begin{minipage}{0.15\textwidth}
%         \centering
%         \includegraphics[width=\linewidth]{figures/BiPegTransfer_noise2.jpg}
%     \end{minipage}
%     % \hfill
%     \hspace{1mm}
%     \begin{minipage}{0.15\textwidth}
%     \centering
%     \includegraphics[width=\linewidth]{figures/GauzeRetrieve_noise2.jpg}
%     \end{minipage}
%     \caption{\TODO{revise the table and caption}perturbing the samples with different scale of noise results in different training results. Three bar charts showing the success rate with models trained with three perturbing settings, including the intermediate models. (a) NeedlePick (b) GauzeRetrieve (c) BiPegTransfer}
%     \label{fig:ablation2}
% \vspace{-3ex}
% \end{figure}

% \begin{figure}[t]
%     \centering
%     \includegraphics[width=\linewidth]{figures/ablation2.png}
%     \caption{
%     We test the performance of our method under different perturbation settings.
%     "$\sigma$" is the mean of the added Gaussian noise and "steps" corresponds to the number of actions being perturbed.}
%     \label{fig:ablation2}
% \vspace{-3ex}
% \end{figure}

\begin{figure}[t]
    \centering
    \includegraphics[width=\linewidth]{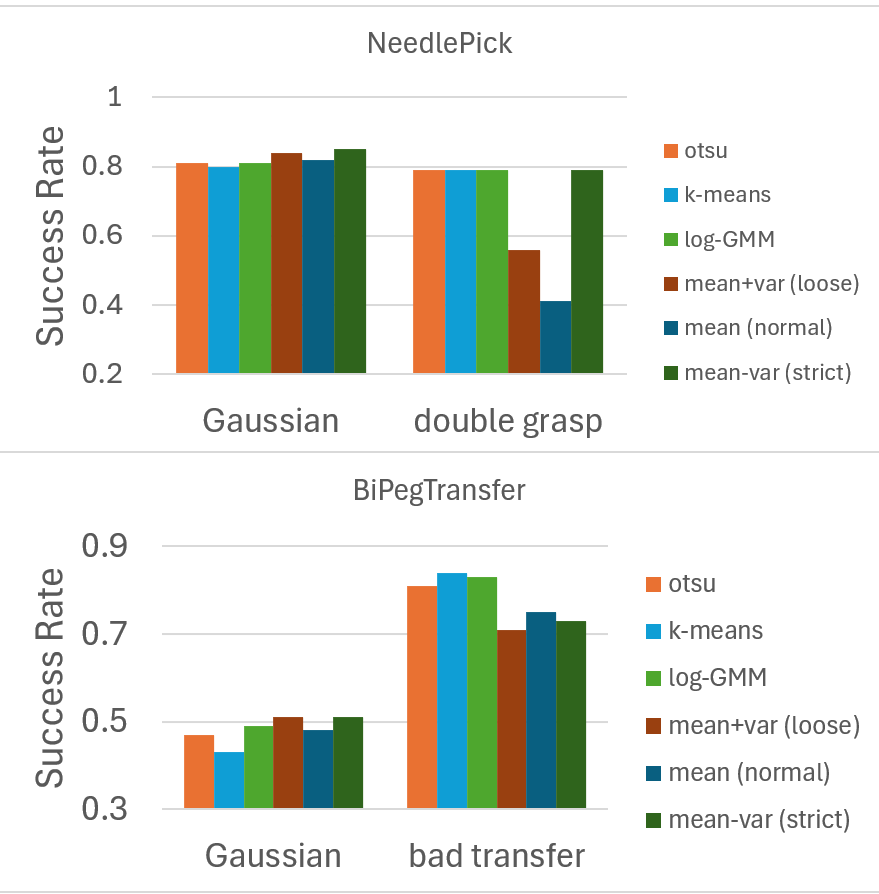}
    \caption{Comparison of six thresholding methods across two surgical tasks (NeedlePick and BiPegTransfer) under two distinct noise configurations (Gaussian step-perturbed and imperfect demonstrations). Results demonstrate that our method maintains robust performance across different threshold selections in both task environments and noise conditions.}
    \label{fig:ablation3}
\vspace{-3ex}
\end{figure}

\begin{figure}[t]
    \centering
    \includegraphics[width=0.95\linewidth]{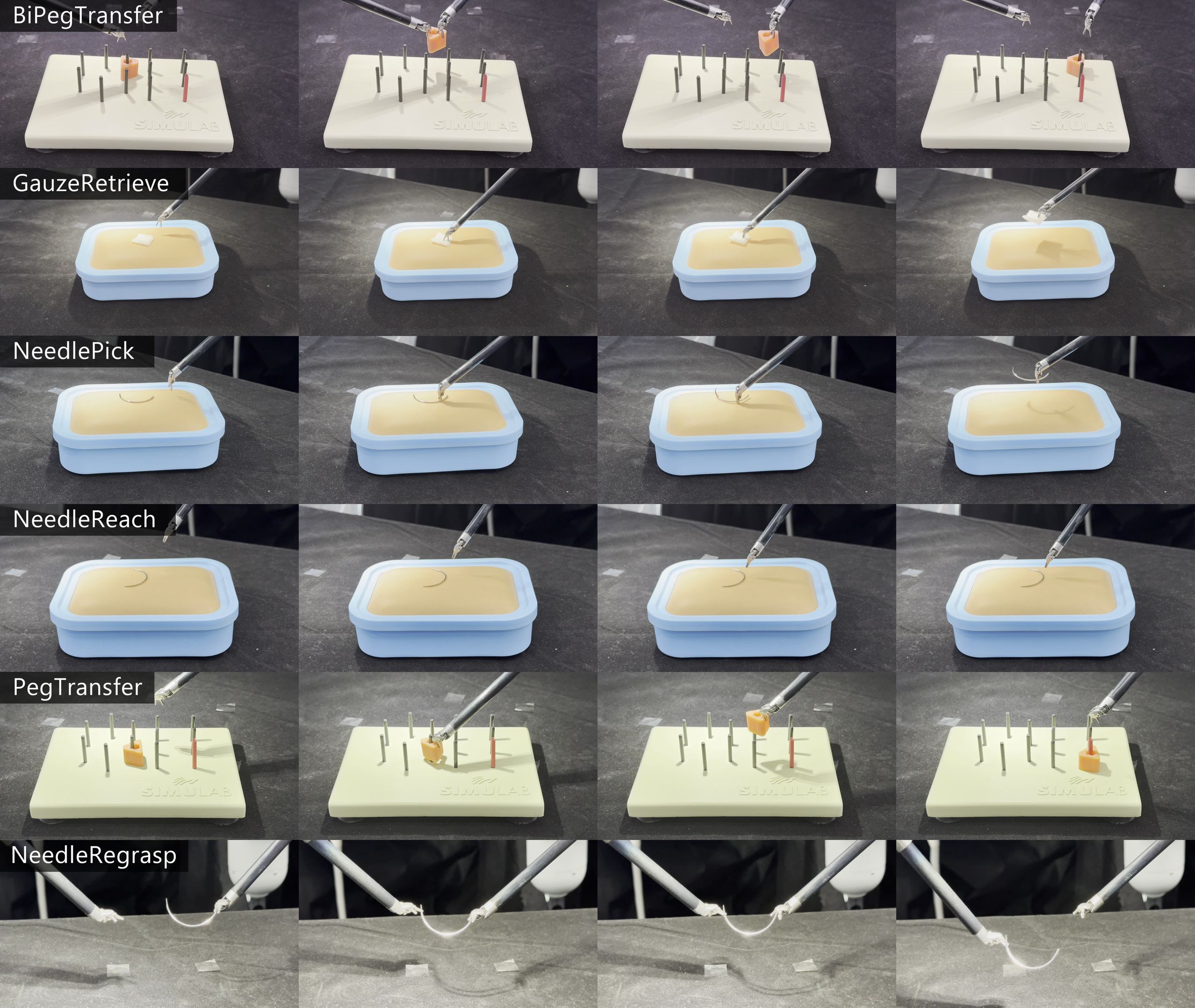}
    \caption{
    Keyframe sequences of task executions on the real robotic platform. 
    Each row corresponds to a surgical task, illustrating its complete workflow. 
    Each column represents the execution state of different tasks at a similar phase (e.g., approach, grasp, transfer, placement).
    }
    \label{fig:real_robot_demo}
    \vspace{-2ex}
\end{figure}

\paragraph{Does our method work with different numbers of clean and perturbed data?}
First, to present the effect of varying numbers of clean demonstrations, a comprehensive evaluation is conducted on the PSM and Bi-PSM tasks.

As shown in \Cref{Tab:different_num_of_demos}, the performance of the proposed DSP method begins to saturate around 75 to 100 episodes, indicating that 100 episodes are generally sufficient for the diffusion policy to capture the task distribution.
% While this demonstrates the ability of the diffusion model to achieve promising performance with limited data,
The stringent success rate requirements of surgical robotics and complexity of real surgical tasks still highlight the necessity fully leveraging limited expert demonstrations.
Therefore, effectively utilizing all available data, including perturbed trajectories, remains critical.

% On one side, the results, shown in \Cref{Tab:different_num_of_demos}, point out another advantage of the diffusion model that it can achieve promising performance with limited data.
% On the other side, considering the significant requirement on high success rate of surgical robot, the results highlight the necessity of enough expert demonstrations using no matter diffusion policy or other learning algorithms.
% Effectively utilizing all available data, including perturbed data, is significant for learning a policy for surgical tasks.

% To demonstrate the effectiveness of our framework, we build 3 mixed datasets with $N$ expert demonstrations and $\tilde{N}$ perturbed episodes, in which $(N,\tilde{N})={(15, 15), (25, 25), (15, 35)}$, respectively.
% The perturbed datasets were generated by introducing synthetic noise into 20\% of the expert actions within each trajectory. We considered three distinct noise types—Gaussian, uniform, and Poisson, and various kind of imperfect demonstration for each task, each designed to simulate different forms of real-world execution imperfections. Detailed specifications of each noise distribution and their parameters are provided in the Appendix. As outlined in the problem setting (see \Cref{sec:method}), these perturbations emulate various accidental or suboptimal operations that may occur during real demonstration collection. 
% Note that the injected noise could lead to task failure in some perturbed trajectories.

To demonstrate the effectiveness of our framework with different constructions of the training set, we construct three mixed datasets containing $N$ clean demonstrations and $\tilde{N}$ perturbed episodes, with $(N,\tilde{N}) \in {(15,15), (25,25), (15,35)}$.
This setup evaluates two aspects of dataset construction, as shown in \Cref{Tab:noisy_ratio_comparison}.
First, we compare the performance of our method under the same ratio of clean and perturbed data, i.e., $(N,\tilde{N})=(25,25)$ vs. $(15,15)$, while scaling up the dataset size. With larger datasets, the diffusion policy can learn a better representation of the data distribution from the additional clean data, thereby reducing the negative impact of perturbed data. Second, we analyze the effect of varying the ratio of clean to perturbed data under the same dataset size, i.e., $(25,25)$ vs. $(15,35)$. It is worth noting that in the last two columns of Table \ref{Tab:noisy_ratio_comparison} for Trajectory-level Perturbation (e.g., $(N,\tilde{N})=(15,35)$), the performance of our method is unexpectedly worse than the baseline for certain tasks like PegTransfer and GauzeRetrieve. This phenomenon occurs because trajectory-level perturbations often contain multi-modal recovery behaviors. Our strict filtering mechanism may inadvertently discard these complex but valid boundary samples, excessively reducing the effective dataset size. Consequently, the diffusion model struggles to generalize, whereas the standard diffusion baseline benefits from the sheer volume of the mixed data despite its noise.

\paragraph{What are the results of using different thresholds for filtering?}
% As stated in \Cref{sec:method}, we use the empirical mean $\hat{\mu}$ of the error $\delta_m$ in \Cref{eq:threshold} as the threshold $\gamma$ to filter the perturbed data within a batch.
% Here, we consider a strict threshold $\hat{\mu}-\hat{\sigma}$:
% \begin{equation}
% \begin{aligned}
%     \gamma &= \hat{\mu}-\hat{\sigma},\\
%     \hat{\mu} = \frac{1}{M} \sum_{m=1}^{M} \delta_m, & \hat{\sigma}^2 = \frac{1}{M-1} \sum_{m=1}^{M} (\delta_m - \hat{\mu})^2,\\
% \end{aligned}
% \end{equation}
% where $\hat{\sigma}^2$ is the empirical variance of the error.
% With this threshold, the diffusion stablizer policy will strictly filter more samples within a batch.
% As shown in \Cref{fig:ablation3},
% while the usage of this threshold results in a slight performance degradation in most environments, our method demonstrates overall robustness and remains largely insensitive to the choice of threshold.
As stated in \Cref{sec:method}, we use a threshold $\gamma$ to filter the perturbed data within a batch.
To further investigate the sensitivity of our method to threshold selection, we introduce several alternatives, which includes a stricter alternative $\hat{\mu}-\hat{\sigma}$ and a loose threshold $\hat{\mu}+\hat{\sigma}$:
\begin{equation}
\begin{aligned}
    \gamma_{strict} &= \hat{\mu}-\hat{\sigma},\\
    \gamma_{loose} &= \hat{\mu}+\hat{\sigma},\\
\end{aligned}
\end{equation}
where $\hat{\sigma}^2$ is the empirical variance of the error. These two thresholds are designed to exclude or accept more of samples in each batch respectively.
Beyond pre-defined thresholds, we also explore adaptive thresholding techniques including Otsu’s method \cite{4310076}, Gaussian Mixture Models (GMM) \cite{melnykov2010finite}, and K-means clustering \cite{hartigan1979algorithm}. Given that the error distribution is heavily concentrated near zero, we apply logarithmic scaling to the errors before computing adaptive thresholds, followed by exponential transformation to revert to the original scale.
The results are shown in \Cref{fig:ablation3}.
% This preprocessing step enhances the discriminative capacity of the adaptive methods in the presence of highly skewed data.

\subsection{Real-World Demonstration}

To validate the practical applicability and sim-to-real transfer capability of our framework, we successfully deployed the policies trained in simulation onto an actual robotic surgical platform, conducting functional validation for all six surgical tasks outlined in our study. As visually documented in \Cref{fig:real_robot_demo}, these real-world demonstrations capture the complete execution workflow of each task, showcasing that our simulation-trained policies generate stable, reliable, and directly transferable motion trajectories for real hardware operation. While constraints inherent to operating precision surgical robotics hardware—including stringent safety protocols and limited availability—precluded large-scale quantitative testing in this initial deployment phase, the observed successful completion of all task objectives provides critical proof-of-concept validation. This demonstration confirms that the behaviors learned in simulation can be effectively transferred to physical systems, thereby verifying the practical viability of our entire pipeline for real-world surgical applications.

\subsection{Discussion and Limitations}
While our synthetic action-level and trajectory-level perturbations are designed to emulate real-world sensor noise and common cognitive errors (e.g., mis-grasping or reaching the wrong goal), it remains an open question whether these non-optimal trajectories fully capture the complexity and biomechanical variations of realistic surgeon behaviors.
% Furthermore, our baseline comparisons primarily evaluate algorithms on their standard training protocols.
% If baseline methods were equipped with similar filtering mechanisms or pre-processed trajectories, their performance might also improve.
Future work will focus on collecting and incorporating actual surgeon-generated suboptimal demonstrations to further validate the realism of our perturbations.
% and on integrating our filtering framework with other baseline algorithms to ensure a more comprehensive comparison.

%%%%%%%%%%%%%%%%% DISCUSSION %%%%%%%%%%%%%%%%%
% As shown in \Cref{fig:ablation3},
% while the usage of this threshold results in a slight performance degradation in most environments, our method demonstrates overall robustness and remains largely insensitive to the choice of threshold.
%%%%%%%%%%%%%%%%%%%%%%%%%%%%%%%%%%%%%%%%%%%%%%%
\section{CONCLUSIONS}
We propose a diffusion-based policy learning framework, called {\ours}, that enables training a diffusion model using a mixture of clean demonstrations and perturbed trajectories.
We showcase the superior performance of our method in the absence of perturbations and its robustness when handling perturbed data.
We hope our approach serves as a step towards applying diffusion policy on surgical tasks, sparking further interest in this direction.
Moreover, we hope our method paves the way for scaling up data in the field of surgical robotics.
% A conclusion section is not required. Although a conclusion may review the main points of the paper, do not replicate the abstract as the conclusion. A conclusion might elaborate on the importance of the work or suggest applications and extensions. 

% \addtolength{\textheight}{-12cm}   % This command serves to balance the column lengths
                                  % on the last page of the document manually. It shortens
                                  % the textheight of the last page by a suitable amount.
                                  % This command does not take effect until the next page
                                  % so it should come on the page before the last. Make
                                  % sure that you do not shorten the textheight too much.

%%%%%%%%%%%%%%%%%%%%%%%%%%%%%%%%%%%%%%%%%%%%%%%%%%%%%%%%%%%%%%%%%%%%%%%%%%%%%%%%

%%%%%%%%%%%%%%%%%%%%%%%%%%%%%%%%%%%%%%%%%%%%%%%%%%%%%%%%%%%%%%%%%%%%%%%%%%%%%%%%

%%%%%%%%%%%%%%%%%%%%%%%%%%%%%%%%%%%%%%%%%%%%%%%%%%%%%%%%%%%%%%%%%%%%%%%%%%%%%%%%
\section*{APPENDIX}

\subsection{Action-level Perturbation Settings}

We considered four types of perturbation noise added to expert actions \(a^*\):

\subsubsection*{Gaussian noise}
A random perturbation \(\epsilon \sim \mathcal{N}(\eta, \sigma)\) with mean \(\eta = 0.05\) and standard deviation \(\sigma = 0.2\), applied independently to each action dimension with probability \(0.5\).

\subsubsection*{Uniform noise}
A perturbation \(\epsilon \sim \mathcal{U}(-0.2, 0.2)\), applied to each dimension with probability \(0.5\).

\subsubsection*{Poisson noise}
A perturbation \(\epsilon \sim \text{Poisson}(\lambda)\) with rate parameter 
\(\lambda = 0.2\). The perturbation is applied with probability \(0.5\).

\subsection{Trajectory-level Perturbation Settings}
This appendix provides a comprehensive classification and description of imperfect demonstration types implemented for each surgical task in simulations. These non-optimal trajectories represent common execution failures that may occur during robotic surgical procedures, providing valuable data for training robust imitation learning algorithms. If not stated, these imperfect demonstrations are implemented by adding way-points for each task during the data generating procedure. In the table, the type numbering of trajectory-level noise corresponds to the order in the following list.

\subsubsection{PSM Tasks}

\paragraph{NeedlePick}  
\textbf{double grasp}: Requires two attempts to successfully secure the needle.  
\textbf{didnt grasp}: Fails to grip the needle despite proper positioning.  
\textbf{wrong goal}: Succeeds in grasping but navigates to an incorrect target.  

\paragraph{NeedleReach}  
\textbf{didn't reach}: Fails to achieve the target coordinates.  
\textbf{bad reach}: first approaches the vicinity, then fine-tuning to the final destination.

\paragraph{PegTransfer}  
\textbf{bad drag}: Suboptimal transportation path with excessive movement via an added way-point.  
\textbf{double grasp}: Requires a secondary attempt to pick up the peg.  
\textbf{miss drop}: Fails to deploy the peg at the target.  

\paragraph{GauzeRetrieve}  
\textbf{double grasp}: Requires two attempts to secure the gauze. 
\textbf{didn't grasp}: Fails to acquire the gauze but continues the remaining movement.  
\textbf{wrong goal}: Retrieves the gauze but transports it to an incorrect destination.  

\subsubsection{Bi-PSM Tasks (Bimanual Manipulation)}

\paragraph{BiPegTransfer}  

\textbf{bad transfer}: Suboptimal coordination between manipulators during object transfer, characterized by awkward kinematic configurations.  

\textbf{double grasp}: Requirement for two grasping attempts by the grasping manipulator during the transfer process.  

\textbf{miss drop}: Failure to accurately deploy the peg at the designated target zone.

\paragraph{NeedleRegrasp}  

\textbf{bad transfer}: Poorly coordinated needle exchange between manipulators, demonstrating suboptimal bi-manual coordination.  

\textbf{didn't grasp}: Failure to successfully transfer needle ownership between manipulators, while the PSMs perform the remaining movements.

\textbf{wrong goal}: Successful needle regrasping followed by navigation to an incorrect target location.

\subsection{Implementation Details of Adaptive Thresholding Methods}

To distinguish clean from noisy samples, we compute thresholds on log-transformed errors (to balance magnitudes) and map them back to the original error scale:
\subsubsection{Otsu's Method}:  A histogram with 40 bins in log-space was built. The threshold was chosen as the value that maximized between-class variance, then mapped back to the original error scale.
\subsubsection{Gaussian Mixture Model (GMM)}: A two-component GMM from scikit-learn was fit to the log-errors, with means initialized at the 10th and 90th percentiles. Training ran up to 500 iterations with covariance regularization (1e-6). The threshold was the intersection where both components had equal posterior probability, then converted back to the original scale.
\subsubsection{K-means Clustering}: Midpoint between two cluster centers computed via scikit-learn’s KMeans.

\section*{ACKNOWLEDGMENT}

This work has been supported in part by the program of National Natural Science Foundation of China (No.62503322), Shanghai Magnolia Funding Pujiang Program (No. 23PJ1404400), and the AI for Science Seed Program of Shanghai Jiao Tong University (project number 2025AI4SQY06).

The work from L. Song and Q. Dou has been supported in part by a grant from the NSFC/RGC Joint Research Scheme sponsored by the Research Grants Council of the Hong Kong Special Administrative Region, China and the National Natural Science Foundation of China (Project No. N CUHK410/23), and in part by a grant from
the Research Grants Council of the Hong Kong Special Administrative Region, China (Project No. 14208424).

% The preferred spelling of the word ÒacknowledgmentÓ in America is without an ÒeÓ after the ÒgÓ. Avoid the stilted expression, ÒOne of us (R. B. G.) thanks . . .Ó  Instead, try ÒR. B. G. thanksÓ. Put sponsor acknowledgments in the unnumbered footnote on the first page.

%%%%%%%%%%%%%%%%%%%%%%%%%%%%%%%%%%%%%%%%%%%%%%%%%%%%%%%%%%%%%%%%%%%%%%%%%%%%%%%%

% References are important to the reader; therefore, each citation must be complete and correct. If at all possible, references should be commonly available publications.

\bibliographystyle{IEEEtran}
\bibliography{mybib}
\end{document}